# Knowledge-Augmented Large Language Model Agents for Explainable Financial Decision-Making

Qingyuan Zhang
Boston University
Boston, USA

Yuxi Wang
Hofstra University
Hempstead, USA

Cancan Hua
University of Southern California
Los Angeles, USA

Yulin Huang
Georgia Institute of Technology
Atlanta, USA

Ning Lyu*
Carnegie Mellon University
Pittsburgh, USA

*Abstract*-This study investigates an explainable reasoning method for financial decision-making based on knowledge-enhanced large language model agents. To address the limitations of traditional financial decision methods that rely on parameterized knowledge, lack factual consistency, and miss reasoning chains, an integrated framework is proposed that combines external knowledge retrieval, semantic representation, and reasoning generation. The method first encodes financial texts and structured data to obtain semantic representations, and then retrieves task-related information from external knowledge bases using similarity computation. Internal representations and external knowledge are combined through weighted fusion, which ensures fluency while improving factual accuracy and completeness of generated content. In the reasoning stage, a multi-head attention mechanism is introduced to construct logical chains, allowing the model to present transparent causal relationships and traceability during generation. Finally, the model jointly optimizes task objectives and explanation consistency objectives, which enhances predictive performance and reasoning interpretability. Experiments on financial text processing and decision tasks show that the method outperforms baseline approaches in accuracy, text generation quality, and factual support, verifying the effectiveness of knowledge enhancement and explainable reasoning. Overall, the proposed approach overcomes the limitations of traditional models in semantic coverage and reasoning transparency, and demonstrates strong practical value in complex financial scenarios.

*Keywords: Knowledge augmentation; large language models; financial decision-making; explainable reasoning*

## I. Introduction

Driven by digital transformation and the wave of intelligence, the financial industry is undergoing profound change. The volume of information in financial markets continues to grow, with data sources spanning news, reports, transaction records, and social media across multiple dimensions. The complexity and timeliness of this information far exceed the past[1]. How to make effective decisions in large-scale, multimodal, and dynamic data environments has become a core challenge for the financial field. Traditional statistical modeling methods have accumulated rich experience in structured data analysis, but they often fall short in handling unstructured text, cross-domain knowledge integration, and dynamic event reasoning[2]. The rise of large language models brings new opportunities for financial decision-making. They demonstrate strong potential in natural language understanding, knowledge representation, and generation, enabling the extraction of key information from massive data and supporting reasoning. However, models that rely solely on parameterized knowledge face limitations in responding to the rapidly changing environment of financial markets, which makes it urgent to introduce external knowledge enhancement mechanisms to ensure completeness and accuracy of information[3].

The introduction of knowledge-enhanced large language model agents brings higher interpretability and transparency to financial decision-making. In practice, financial decision-making requires not only correctness of results but also traceability of reasoning paths and verifiability of logical chains. This is because financial activities involve significant economic interests and social impact, and any incorrect or non-explainable judgment may trigger systemic risks. By integrating external knowledge, agents can invoke authoritative data sources and professional knowledge bases during reasoning, ensuring the rationality and credibility of outputs. At the same time, explainable reasoning mechanisms help researchers and decision-makers understand the basis of model conclusions, strengthening their trust in the system. This not only enhances the application value of artificial intelligence in the financial sector but also provides new solutions for regulatory review and compliance requirements.

In the context of financial decision-making, knowledge-enhanced explainable reasoning holds multiple meanings. First, it improves the robustness of decision-making and helps address high-noise environments and uncertainty in financial markets. By performing semantic-level restructuring and cross-validation of information, the model can better resist misinformation or bias from distorted data. Second, it facilitates the discovery of potential causal relationships, revealing the logical structure behind market dynamics rather than only surface correlations. This capability is particularly critical for risk identification, trend forecasting, and event-driven analysis in financial scenarios. Third, the combination of knowledge enhancement and explainable reasoning supports cross-domain information integration, allowing financial decisions to take into account macroeconomic trends, industry dynamics, and individual behaviors, thereby forming multi-level and multi-dimensional comprehensive judgments[4].

II. RELATED WORK

Related work on knowledge-enhanced large language models and explainable decision-making can be broadly grouped into retrieval-augmented reasoning, knowledge-structured representation and explainability, advanced anomaly and risk modeling architectures, and agent-based learning and decision frameworks.

Retrieval-augmented large language models (LLMs) provide an important foundation for combining internal parametric knowledge with external information sources. Self-reflective retrieval-augmented methods jointly learn to retrieve, generate, and critique, forming a closed loop in which the model iteratively improves its own retrieval and generation behavior through self-evaluation and refinement [5]. Fusion-based retrieval-augmented generation frameworks further design multi-branch retrieval and fusion mechanisms, allowing different retrieval channels and evidence sources to be integrated into a unified semantic representation before prediction, which enhances factual completeness and robustness in complex reasoning scenarios [6]. Information-constrained retrieval with LLM agents extends this line by organizing retrieval, filtering, and reasoning into multi-stage pipelines under explicit constraints on information usage, ensuring that the model remains efficient, controllable, and aligned with task requirements [7]. For financial text processing, sparse retrieval combined with deep language modeling has been explored for robust fact verification, where retrieval components and language models are jointly optimized to improve the alignment between retrieved evidence and generated or predicted outputs [8]. Benchmark efforts for financial question answering define challenging evaluation suites that stress knowledge understanding, numerical reasoning, and factual consistency, pushing LLM-based systems toward more reliable decision-making behaviors [9]. Methodologically, these works demonstrate how retrieval modules, fusion strategies, self-critique, and task-specific benchmarks can be combined with large language models to build knowledge-aware reasoning systems. The framework in this study follows the same spirit by encoding financial texts and structured data, retrieving task-related knowledge, and fusing internal and external representations through weighted mechanisms to enhance factual support and decision quality.

Another line of research emphasizes structured representation, knowledge integration, and explainability in neural language models. Methods that integrate knowledge graph reasoning with pretrained language models inject structured relational information into the representation space through joint optimization of reasoning operators and language encoders, thereby enriching the semantic structure captured by the model and improving its ability to perform structured inference [10]. Structure-aware attention mechanisms combined with auxiliary knowledge representations provide a way to guide attention weights using relational and structural cues, which enhances both performance and interpretability by making the contribution of different entities and relations more transparent [11]. From a broader perspective, systematic analyses of explainable artificial intelligence in finance summarize major paradigms of explanation, common model families, and evaluation dimensions, highlighting the need to balance predictive performance with transparency, traceability, and regulatory requirements [12]. In parallel, deep learning and natural language processing methods for unified summarization and structuring of complex documents show how unstructured text can be transformed into more organized representations that support downstream reasoning and decision tasks [13]. These methodological ideas—knowledge graph integration, structure-aware attention, systematic explainability, and text-to-structure transformation—are closely related to the design in this paper. The proposed knowledge-enhanced large language model agent constructs semantic representations from financial texts and structured data, fuses them with retrieved knowledge through weighted mechanisms, and introduces multi-head attention to explicitly model reasoning chains and logical dependencies, thereby improving both decision performance and explanation quality.

Advanced modeling techniques for anomaly detection and risk discrimination provide additional components relevant to robust financial decision-making. Sequence modeling with refined state-space architectures has been used to discriminate subtle patterns in transactional data, demonstrating that improved sequence models can better capture long-range dependencies and regime shifts in complex temporal signals [14]. Metric-learning-based frameworks with Siamese architectures have been proposed for risk discrimination tasks, where the model learns to compare pairs of instances in a shared representation space and optimize distance-based objectives for anomaly identification and risk assessment [15]. Few-shot anomaly segmentation via dictionary lookup introduces a framework that builds class-generalizable dictionaries of normal and abnormal patterns and leverages them to detect anomalies with limited supervision, illustrating how dictionary-based representation and matching can provide robust detection capabilities in data-scarce regimes [16]. These techniques share a focus on discriminative representation learning, robust sequence modeling, and anomaly-sensitive decision criteria. The method proposed in this study is complementary: it focuses on explainable reasoning and knowledge integration in financial decision tasks, while also emphasizing factual support and risk-awareness, and it can benefit from advanced sequence and metric-learning components when modeling temporal financial signals or complex risk patterns. Agent-based and reinforcement learning paradigms highlight adaptive, coordinated decision-making in dynamic environments [17,18]. Building on these lines, our framework uses a knowledge-enhanced LLM agent that encodes financial texts and structured data, retrieves task-relevant knowledge, fuses it via weighted attention, and jointly optimizes prediction and explanation to deliver fact-supported, interpretable financial decisions.

## III. METHOD

In this study, we apply a knowledge-enhanced large language model agent designed for explainable financial decision-making, directly incorporating several state-of-the-art techniques. During input encoding, we adopt the contextual representation and attention strategies proposed by Wu and Pan [19] to convert both financial texts and structured data into rich semantic embeddings. For external knowledge retrieval, our framework utilizes the selective knowledge injection approach with adapter modules introduced by Zheng et al. [20], enabling the agent to draw on external knowledge bases and inject domain-specific information where needed. At the semantic fusion stage, we integrate Xue's [21] dynamic structured gating mechanism, which efficiently aligns and merges internal representations with retrieved knowledge to maintain factual consistency and parameter efficiency. This multi-source fusion provides a strong foundation for the final reasoning generation stage, where the agent constructs interpretable reasoning chains for transparent and explainable outputs. The overall model architecture is depicted in Figure 1.

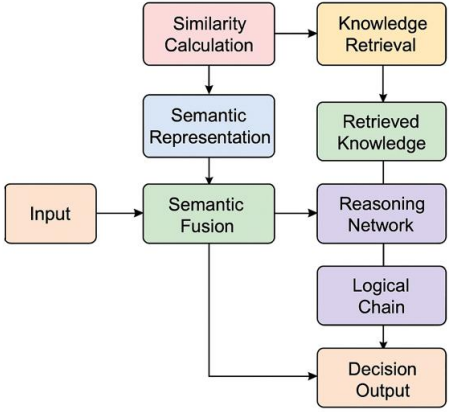

Figure 1. Overall model architecture

Suppose the original input sequence is $x = \{x_1, x_2, ..., x_n\}$, which is passed through the encoder to obtain the context representation matrix:

$$H = Encoder(x) = [h_1, h_2, ..., h_n] \quad (1)$$

Where $h_i$ represents the vectorized representation of the i-th element in the input sequence.

After obtaining the internal semantic representation, the model needs to enhance its reasoning capabilities through external knowledge. To achieve dynamic knowledge retrieval, the input semantic vector is first normalized with the candidate vectors in the external knowledge base, and the first k knowledge fragments retrieved are used as supplementary information [22]. The similarity is measured using the cosine distance, which is defined as follows:

$$sim(h, k_j) = \frac{h \cdot k_j}{\|h\| \|k_j\|} \quad (2)$$

Where $h$ represents the semantic representation of the input and $k_j$ is the candidate knowledge vector in the external knowledge base. Subsequently, the internal and external representations are fused through a weighted mechanism to obtain the knowledge-enhanced representation:

$$z = ah + (1-a)\sum_{j=1}^{k} w_j k_j \quad (3)$$

Among them, $a$ is the balance coefficient and $w_j$ is normalized according to the similarity distribution. The fused representation z serves as the input to the attention layer. The token's internal semantic vector $h$ and the retrieved external knowledge vectors $\{k_j\}$ jointly influence the computation of attention weights.

During the inference phase, the agent not only needs to make predictions using the enhanced representation but also needs to generate interpretable logical chains. To this end, an inference network based on the attention mechanism is constructed to interact with information from different sources. Let the enhanced representation be z, and the inference process is updated through the self-attention mechanism as follows:

$$Z' = \text{softmax}\left(\frac{QK^T}{\sqrt{d}}\right)V \quad (4)$$

Where $Q, K, V$ represents the query, key, and value matrices, and represents the vector dimension. Q,K,V can be derived from the fused representation through learned linear projections: $Q=W_Q z$   $K=W_K z$ , $V=W_V z$ . This design allows external knowledge to reshape the similarity structure used by attention: tokens that share relevant knowledge elements are more strongly aligned in the $QK^T$ space, and tokens influenced by conflicting or inconsistent knowledge receive diminished attention weights. The attention mechanism does not only aggregate contextual information from the input sequence but also dynamically redistributes focus based on the factual evidence retrieved from the knowledge base. This mechanism ensures the dynamic interaction and logical consistency of each part of the inference chain.

Finally, to achieve interpretable reasoning, this study introduces logical constraints and traceability goals. Specifically, when the model generates the final decision output, it needs to minimize both semantic error and interpretation bias in the loss function. The overall optimization goal is defined as:

$$L = L_{task} + \lambda L_{explain} \quad (5)$$

Here, $L_{task}$ represents the semantic prediction error of the financial task, $L_{explain}$ represents the constraint on the interpretation chain and knowledge consistency, and $\lambda$ is the trade-off coefficient. Through this optimization process, the intelligent agent can provide explainable reasoning while ensuring predictive performance, thus meeting the transparency and credibility requirements of financial decision-making.

IV. EXPERIMENTAL RESULTS

*A. Dataset*

The dataset used in this study is the FiQA Dataset, which consists of text corpora from the financial domain. It mainly covers financial news, company announcements, and user discussions in financial communities. The dataset was designed to support financial question answering and opinion mining tasks, and it has strong representativeness in terms of corpus scale and content coverage. It contains news headlines, article bodies, and question-answer pairs related to financial events, which reflect the diversity and complexity of semantic expressions in the financial domain.

All texts in this dataset come from real financial contexts, including topics such as stocks, bonds, currencies, and macroeconomics. The data is highly time-sensitive and noisy in semantics. These features are consistent with the actual operating mechanisms of financial markets, which makes the dataset a reliable basis for financial decision-making tasks. The question-answer structure and sentiment annotations help models consider both factual information and subjective expressions, thereby supporting more complex reasoning and explanation generation. The choice of the FiQA Dataset lies in its wide applicability and scalability in financial text processing tasks. Compared with general corpora, this dataset is closer to the language style and knowledge requirements of the financial field, which enables effective validation of the advantages of knowledge enhancement and explainable reasoning in professional scenarios. In addition, the openness and standardized annotation of the dataset ensure the reproducibility and comparability of research results, providing a solid foundation for further exploration in the direction of intelligent financial decision-making.

*B. Experimental Results*

This paper first gives the comparative experimental results, as shown in Table 1.

Table1. This paper first presents the comparative experimental results

| Model | Accuracy | ROUGE | BLEU | FactScore |
|---|---|---|---|---|
| Expel[23] | 71.8 | 34.2 | 21.5 | 0.62 |
| Aios[24] | 74.6 | 36.8 | 23.1 | 0.65 |
| Agent-safetybench[25] | 76.3 | 38.1 | 24.0 | 0.67 |
| Agentlite[26] | 77.5 | 39.5 | 25.2 | 0.70 |
| Ours | 82.7 | 44.8 | 29.6 | 0.79 |

From Table 1, existing methods (e.g., Expel, Aios, Agent-safetybench, Agentlite) achieve moderate Accuracy and ROUGE but remain weak on FactScore, indicating limited factual grounding in financial decisions. In contrast, the proposed method achieves the best results on Accuracy, ROUGE, BLEU, and a much higher FactScore of 0.79, showing both better generation quality and stronger knowledge consistency. The sensitivity of ROUGE to batch size is further analyzed in Figure 2.

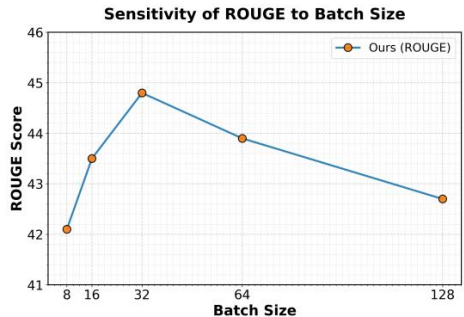

Figure 2. Batch size sensitivity experiment on ROUGE

From Figure 2, batch size clearly affects ROUGE. Small batches (8, 16) yield lower ROUGE, suggesting unstable semantic modeling and insufficient integration of external knowledge. ROUGE peaks at batch size 32, indicating the best balance between stable gradients, contextual sensitivity, and knowledge utilization for dense financial texts. Larger batches (64, 128) reduce ROUGE, likely due to over-smoothed updates that blur fine-grained semantics and weaken explainable reasoning. This highlights batch size as a key hyperparameter for credible, knowledge-enhanced financial generation. A separate sensitivity experiment on data noise ratio and ROUGE is reported in Figure 3.

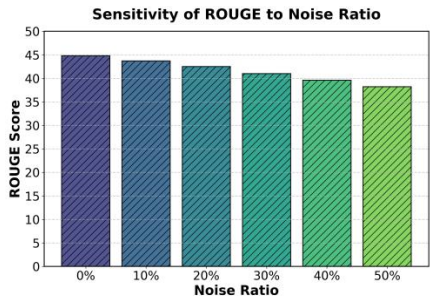

Figure 3. Sensitivity experiment of data noise ratio to ROUGE

From Figure 3, ROUGE decreases monotonically as the noise ratio increases, with the best generation quality at 0% noise, where knowledge enhancement and semantic modeling work most effectively. At 10%–20% noise, ROUGE declines mildly, indicating that the agent remains usable but less stable under imperfect data. When noise reaches 30%–50%, the drop becomes obvious, as irrelevant or incorrect information disrupts knowledge utilization and reasoning coherence, though the model still retains some robustness. These results highlight the critical importance of data quality and noise control for reliable, explainable financial decision-making.

V. CONCLUSION

This study conducts a systematic investigation of knowledge-enhanced large language model agents for explainable reasoning in financial decision-making. The aim is to address the limitations of traditional methods in factual consistency, transparency, and knowledge coverage. By introducing external knowledge and semantic fusion mechanisms, the model achieves traceable logical chains while maintaining language fluency, thereby enhancing reliability and interpretability in financial decision contexts. The results show that the proposed method not only demonstrates advantages in semantic understanding and text generation but also achieves good performance in the completeness and consistency of reasoning chains. These findings provide a solid theoretical and methodological foundation for building intelligent decision-making systems.

In the financial industry, the core demand of intelligent decision-making lies not only in the correctness of results but also in the transparency and verifiability of the reasoning process. The proposed method combines external knowledge bases with internal representations, effectively improving the factual support of the model. As a result, agents can generate outputs with logical consistency and auditability when dealing with complex financial texts and events. This capability is of significant value for key applications such as risk management, compliance review, and market analysis. It helps alleviate information asymmetry, enhances the stability of financial systems, and provides technical support for regulation.

In addition, the explainable reasoning framework proposed in this study has important implications for broader intelligent decision-making domains. By emphasizing knowledge enhancement and logical chains in the modeling process, this research offers a new paradigm for artificial intelligence applications in high-risk and high-complexity environments. This approach promotes trustworthy applications of artificial intelligence in the financial sector and also provides a reference path for knowledge-driven intelligent systems across domains. The value of this research lies in overcoming the limitations of "black-box" models by integrating the strong generative capacity of language models with the demand for knowledge transparency. It offers guidance for developing agents who are more responsible and practical in related fields.